\documentclass[letterpaper, 10 pt, conference]{ieeeconf}  % Comment this line out if you need a4paper
\IEEEoverridecommandlockouts                              % This command is only needed if 
\usepackage{graphicx}
\usepackage{amsmath}
\usepackage{amssymb}
\usepackage{xcolor}
\usepackage{booktabs}
\usepackage{makecell}
\usepackage{multirow}
\usepackage{tabularx}
\usepackage[misc]{ifsym}
\title{\LARGE \bf
Industrial-Grade Sensor Simulation via Gaussian Splatting: \\ A Modular Framework for Scalable Editing and Full-Stack Validation
}

\author{Xianming Zeng$^\dag$, Sicong Du$^\dag$, Qifeng Chen$^\dag$, Lizhe Liu, Haoyu Shu, Jiaxuan Gao, Jiarun Liu, Jiulong Xu, \\ Jianyun Xu, Mingxia Chen, Yiru Zhao, Peng Chen, Yapeng Xue, Chunming Zhao, Sheng Yang\textsuperscript{\Letter}, Qiang Li% <-this % stops a space
\thanks{*All authors are with the Unmanned Vehicle Department of CaiNiao Inc., Alibaba Group.
$\dag$ contributed equally to this work, and \Letter \ denotes the corresponding author and project leader. {Email: \texttt{\{zengxianming.zxm, dusicong.dsc, chenqifeng.cqf,shengyang\}@cainiao.com}}
}%
}

\begin{document}

\maketitle
\thispagestyle{empty}
\pagestyle{empty}

\begin{abstract}

Sensor simulation is pivotal for scalable validation of autonomous driving systems, yet existing Neural Radiance Fields (NeRF) based methods face applicability and efficiency challenges in industrial workflows. This paper introduces a Gaussian Splatting (GS) based system to address these challenges: We first break down sensor simulator components and analyze the possible advantages of GS over NeRF. Then in practice, we refactor three crucial components through GS, to leverage its explicit scene representation and real-time rendering: (1) choosing the 2D neural Gaussian representation for physics-compliant scene and sensor modeling, (2) proposing a scene editing pipeline to leverage Gaussian primitives library for data augmentation, and (3) coupling a controllable diffusion model for scene expansion and harmonization. We implement this framework on a proprietary autonomous driving dataset supporting cameras and LiDAR sensors. We demonstrate through ablation studies that our approach reduces frame-wise simulation latency, achieves better geometric and photometric consistency, and enables interpretable explicit scene editing and expansion. Furthermore, we showcase how integrating such a GS-based sensor simulator with traffic and dynamic simulators enables full-stack testing of end-to-end autonomy algorithms. Our work provides both algorithmic insights and practical validation, establishing GS as a cornerstone for industrial-grade sensor simulation.

\end{abstract}
\section{Introduction}
\label{sec:intro}

Modern autonomous driving systems require exhaustive testing across billions of Operational Design Domains (ODDs), making the physically accurate sensor simulation indispensable. While Neural Radiance Fields (NeRF)~\cite{mildenhall2021nerf} have advanced photorealistic rendering, their implicit geometry representation and high computational demands limit scalability in industrial workflows. Recent advances in \emph{Gaussian Splatting (GS)}~\cite{Kerbl20233dgs,wu20243dgssurvey} offer explicit scene parameterization and real-time rendering, presenting an opportunity to refactor sensor simulation pipelines~\cite{yang2023unisim}.

Three key functional requirements on sensor simulators can be achieved through GS and its accompanying approaches: The first and the cornerstone requirement is to \emph{maintain realism via affordable expense}. In decades, photo-realism has been ensured by exquisite synthetic scene modeling, which requires heavy model and art costs but is still distinguishable from real sensor data~\cite{dosovitskiy2017carla}. NeRF-based simulators~\cite{yang2023unisim} choose to reconstruct from raw scans and thus greatly reduce the scene modeling cost. GS inherits the view-direction-aware rendering of NeRF and further improves the rendering speed, which additionally relieves the computing resources when sensor frames are rendered from these reconstructed scenes.

Secondly, sensor simulators are also extensively leveraged for the data augmentation of perception tasks. They often integrate scene editing capabilities to output realistic scans when scene objects are removed or injected. This reduces the cost of designated data collection and sensor frame labeling. Specifically, in this application, the requirement for sensor simulators is to \emph{produce paired sensor frames along with their corresponding ground-truth labels}. While synthetic scenes inherently have this capability, reconstruction-based methods require de-compositing lighting and shading parameters to maintain harmony after editing. NeRF, as an implicit representation, is comparably complicated for performing object decoupling and scene editing. In contrast, GS can facilitate the scene editing procedure through explicit drag-and-place operations based on scene contexts~\cite{wang2024gscream} and standalone object databases~\cite{du20243drealcar}.

Thirdly, when autonomous vehicles are shipped from definite areas to public roads, they occasionally obtain valuable driving clips for reviewing and iterating algorithms. Hence, the industries use a crowd-sourced data collection paradigm to obtain these clips passively. In this passive scheme, reconstruction methods receive single-pass scans and result in incomplete scene representation, degrading their effectiveness on free-viewpoint sensor simulation. This makes it difficult to perform end-to-end regression testing. For example, when using real-world driving cases to verify a detour rather than a slow-down decision, physically realistic sensor frames from laterally shifted viewpoints are required for regressing perception and end-to-end algorithms. Hence, reconstruction methods should also have \emph{scene expansion ability while keeping physical compliance}. As Diffusion Models~\cite{rombach2022high} have recently emerged as a powerful tool for generating appearance-consistent videos, coupling these methods into a reconstruction pipeline~\cite{fu2024limsim++,Zhao2024drivedreamer4d,fan2024freesim} supports reasonable and physically coherent free-viewpoint rendering.

Regarding all three requirements, we propose a GS-based sensor simulator to enhance the functionality. Specifically, we address the following core components: (1) A chunk-based 2D neural Gaussian representation for paralleled large-scale reconstruction and real-time rendering. (2) An explicit editing approach for placing specific objects into the scene, obtaining a harmonious appearance with ground truth labels. (3) A tightly coupled scheme for reasonably leveraging controllable Diffusion Models~\cite{rombach2022high} to expand the static scene, enabling extrapolated novel view synthesis on single-pass scans. Since L4 autonomy still requires LiDAR sensors, we support the workflow of multiple cameras and LiDAR sensors on these components, covering their fisheye lens distortion and ray-drop characteristics. Based on such a sensor simulator, we demonstrate its effectiveness in (1) improving multiple perception tasks in a data augmentation fashion and (2) regressing end-to-end algorithms on specific scenarios when cooperating with traffic and dynamic simulators. In conclusion, our paper includes the following key contributions:

\begin{itemize}
\item We break down crucial components of a sensor simulator, discussing their necessity and possible profits when refactored by GS.
\item We perform an industrial-grade practice on several crucial components to bring up the GS-based sensor simulator, providing ablation studies on the effects of choosing different algorithms and strategies on our proprietary scans.
\item We showcase how such a sensor simulator improves developing driving autonomy by facilitating both the standalone perception module and the end-to-end driving agent.
\end{itemize}

\section{Systematic Components and Related Works}
\label{sec:system}

\begin{figure}[ht] 
\includegraphics[width=\linewidth]{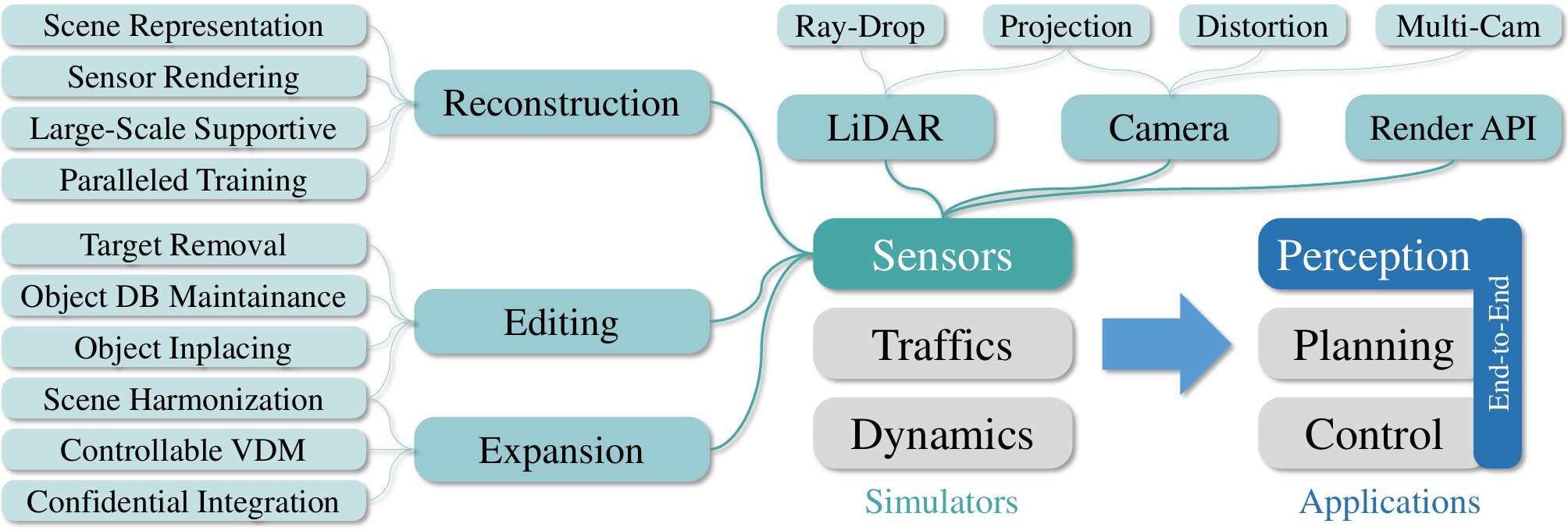}
\caption{Key refactorable components (green) and applications (blue) of the sensor simulator where GS are leveraged to improve performance.}
\label{fig:design} 
\end{figure}

We refer readers to a recent survey~\cite{li2024choose} that regards sensor simulators as supportive of sensory data processing and onboard perception tasks. To meet the requirements listed in Sec.~\ref{sec:intro}, we break down core-related components for upgrading sensor simulators by GS as shown in Fig.~\ref{fig:design}, and review recent systems and approaches accordingly.

\textbf{Scene Representations.}
Several simulators~\cite{dosovitskiy2017carla} use OpenGL or Unreal rendering engine to generate photo-realistic sensor frames, which requires exquisite virtual scene modeling, texture mapping, and render parameters adjusting. To cost-down the scene manufacturing, reconstruction methods~\cite{mildenhall2021nerf,Kerbl20233dgs,wu20243dgssurvey} choose to recover 3D geometry, texture, and rendering parameters from sensor scans, making the simulation results closer to the real scans. Through the last decade, reconstruction methods mainly experienced three types of scene representations:
(1) Direct representations such as SDF volume~\cite{dai2017bundlefusion} and Surfels~\cite{whelan2015elasticfusion} explicitly store RGB attributes in the 3D Euclidean space, reaching real-time efficiency on both reconstruction and rendering, but their photo-realism is limited -- especially for outdoor scenes suffering complicated lighting conditions and variated surface materials.
(2) NeRF representations~\cite{mildenhall2021nerf} enhance photo-realism by marching rasterized rays and query along their intersected primitives for volume rendering, boosting several sensor simulators~\cite{yang2023unisim,ljungbergh2024neuroncap,tonderski2024neurad} to leverage NeRF for scene manufacturing. However, these methods suffer higher computational costs and are thus inefficient enough for digesting crowd-sourced data.
(3) 3D Gaussian representations~\cite{Kerbl20233dgs} follow the ray-marching idea of NeRF, but reversely splat intersected primitives onto rasterized sensor frames for acceleration, where Spherical Harmonics are leveraged for view-direction-aware appearance modeling. 2DGS~\cite{huang20242dgs} replaces 3D Gaussians with 2D Gaussians to accurately perform ray-splat intersection, employing a low-pass filter to avoid degenerated line projection.
Meanwhile, some GS variants~\cite{lu2024scaffold} choose to use neural Gaussians instead of Spherical Harmonics, reaching higher quality with an acceptable trade-off in rendering efficiency. Compared to NeRF, these GS methods achieve real-time rendering speed.
Our method compares and chooses the 2D neural Gaussian representation to refactor sensor simulation, and further integrates parallel training strategies~\cite{zhao2024scaling} to reduce the atomic training time for each sequence.

\textbf{Sensor Characteristics.} High-level autonomy requires multiple and various sensors. As the two most prevalent exteroceptive sensors equipped on vehicles, LiDAR and cameras are crucial cues for perceiving surroundings. Reconstructing or rendering based on these sensors has several sensor-specific issues to address, including the ray-drop pattern of LiDAR frames~\cite{jiang2025gslidar,chen2024lidar}, and the fisheye lens distortion of camera frames~\cite{huang2024ogs}. 
Our system considers these characteristics through a unified projection scheme and maintains a unified representation for both LiDAR Gaussians and camera Gaussians. This forms a generic pattern for multi-modal scene manipulation.

% Our system integrates previous approaches, and further merges reconstructed assets from multiple cameras into one Gaussian representation for improving data efficiency and rendering quality.

\begin{figure*}[t] 
\includegraphics[width=\linewidth]{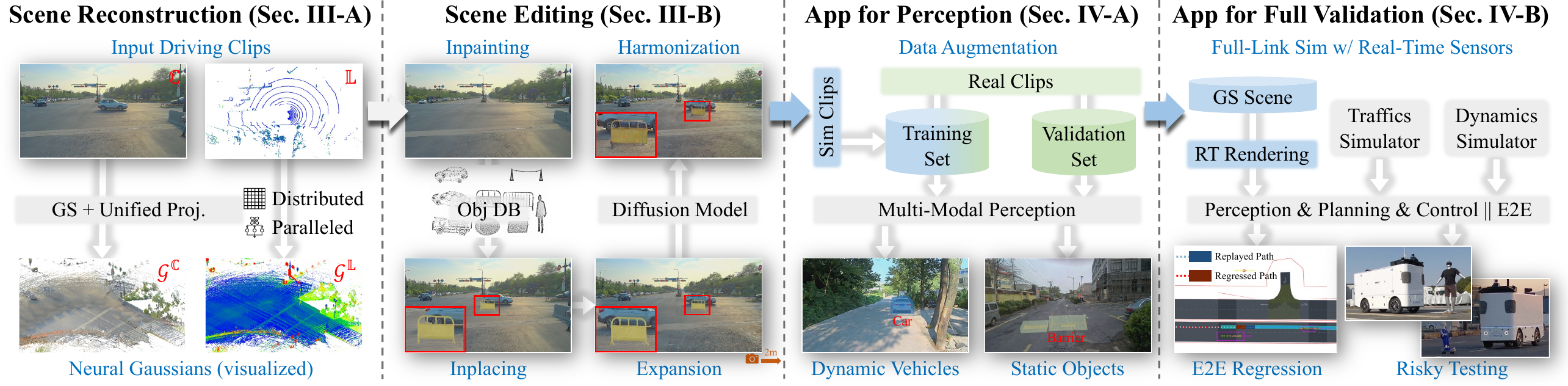}
\caption{The workflow of our proposed GS-based sensor simulator. Given an input driving clip, we first reconstruct various sensors into their corresponding neural Gaussians, where editing, expansion, and harmonization operations can be applied to manipulate all reconstructed Gaussians synchronously. In application phases, it can either be used to enlarge the training set of multi-type perception tasks, or link with other simulators for end-to-end testing.}
\label{fig:workflow} 
\end{figure*}

\textbf{Scene Editing.} Most editing approaches for NeRF representations deform~\cite{yuan2022nerf} or distill~\cite{zhuang2023dreameditor} explicit triangular mesh for responding to user edits. Some other methods design shape prior~\cite{yang2023unisim} or extended neural feature fields~\cite{tonderski2024neurad} to refine and decouple foreground objects during learning. In contrast, GS-based scene editing approaches are relatively simple and typically contain two learning-free stages: Inpainting surroundings to keep the harmony when undesired foreground objects are subtracted~\cite{wang2024gscream,liu2024infusion}, and placing object-level assets -- either individually scanned~\cite{du20243drealcar} or CAD modeled~\cite{deitke2023objaverse} -- to meet the user specification. Meanwhile, relighting approaches~\cite{shi2023gir,gao2024relightable} can be incorporated as optional choices for further improving rendering quality. Through our practice, we demonstrate the effects of drag-and-place operations on Gaussians for improving perception algorithms, proving the simplicity of such an explicit representation.

\textbf{Scene Expansion and Harmonization.} Scene Expansion requires generative AI~\cite{zhu2024sora}. While 3D generative AI~\cite{TripoSR2024} is trying to break the scaling law, 2D Diffusion Models~\cite{rombach2022high,agarwal2025cosmos} as mature substitutes have provided an impressive ability to directly generate high-fidelity and temporal consistent videos, where users can specify particular weather and lighting conditions through text prompts for customization. Since these methods are computationally expensive and indirect for tuning specific details of scenes, several concurrent works~\cite{fan2024freesim,Zhao2024drivedreamer4d} attempt to integrate generated scans into the scene, and reciprocally use scene rendered results to guide the video generation. This strategy achieves a mutual benefit of leveraging scene representations to ensure a reasonable physics for content generation, and expanding the range of reconstructed scenes upon crowd-sourced incomplete scans. Our implementation follows this trend and additionally proves its effectiveness in improving perception tasks. We also found these models effective for partially substituting the aforementioned relighting methods to enhance the harmony of object insertion.

\textbf{Sensor Simulator for Perception.} Given the assurance of realism and a small sim-to-real gap, simulated scans can enrich the acquired data with low scanning and annotation costs, facilitating the development of perception~\cite{Yang2023BEVFormerVA,liao2024lane}. Many GS approaches prove their photo-realism by comparing with real scans, but only a few~\cite{zhang2025mapgs} attempt to feed simulated frames to the training process of models and verify whether perception performance is improved. We prove our effectiveness of leveraging simulated scans on training both static and dynamic object detection tasks, gaining precision on real-scan validation sets. 

\textbf{Sensor Simulator for Full-Stack Validation.} Planning and control, as downstream algorithms of perception, determine and relay operating instructions to the chassis. These algorithms also have their corresponding simulators~\cite{li2024choose}, namely traffic and dynamic simulators. Hence, if we reach both sufficient quality for perception and real-time rendering efficiency for sensor frame deduction, we can launch all algorithms from sensors to motion in a simulated world for a full-stack validation -- the closed-loop simulation. As recent advances in onboard algorithms are chasing an end-to-end fashion~\cite{hu2023uniad,chen2024endtoend}, close-loop simulators with qualified sensor simulation are regarded as a core infrastructure for safe testing~\cite{hu2023simulation}. In this paper, we regress the influence of our sim-to-real gap and showcase our close-loop simulation ability on several high-risk scenarios.

\section{Components Implementation and Ablation}
\label{sec:method}

Fig.~\ref{fig:workflow} illustrates our workflow connecting listed components in Fig.~\ref{fig:design} for bringing-up and applying a GS-based sensor simulator. Besides, we also provide ablation studies on several key components and strategies in Sec.~\ref{sec:method:ablations}.

\subsection{Gaussian-Based Scene Representation}
\label{sec:method:recon}

\textbf{Scene Reconstruction.} Given a driving clip containing sequential and multiple cameras and LiDAR sensor frames, our method chooses to separately reconstruct camera Gaussian scenes $\mathcal{G}^\mathbb{C}$ and one LiDAR Gaussian scene $\mathcal{G}^\mathbb{L}$, respectively for two types of sensors. For both scenes, we use embedded anchors $a \in \mathcal{G}$ for representation organized as Scaffold-GS~\cite{lu2024scaffold}, and Neural Scene Graphs (NSG)~\cite{ost2021neural} to handle dynamic objects as Street Gaussians~\cite{yan2024street}. The final 2DGS~\cite{huang20242dgs} attributes $\{\boldsymbol{\mu}, \mathbf{R}, \mathbf{S}, \rho, r, \alpha\}$ for each neural Gaussian in $\mathcal{G}^\mathbb{C}$ and $\mathcal{G}^\mathbb{L}$ are inferred by passing the anchors features through the corresponding Multi-Layer Perceptron (MLP) network~\cite{chen2024lidar}.

\textbf{Unified Projection Scheme.}
We follow the definition of 2DGS~\cite{huang20242dgs} to define each 2D Gaussian $(u,v)$ in a local tangent plane, whose 3D position in the world coordinate is calculated as:
\begin{equation}
\mathbf{P}(u,v)=\mathbf{W}\mathbf{H}(u,v,1)^\top, \ \mathbf{H} \ \triangleq
\begin{bmatrix}
s_u\boldsymbol{t}_u & s_v\boldsymbol{t}_v & \boldsymbol{\mu} \\
0 & 0 & 1
\end{bmatrix} \in \mathbb{R}^{4\times3},
\end{equation}
where $\boldsymbol{t}_u , \boldsymbol{t}_v \in \mathbb{R}^{1\times3}$ are two rotation component (equivalent to the tangential direction of the Gaussian plane), $s_u$ and $s_v$ are scaling component, $\boldsymbol{\mu} \in \mathbb{R}^{1\times 3}$ is center of the Gaussian plane, $\mathbf{W} \in \mathbb{R}^{4\times4}$ is the model view matrix, and $\mathbf{P} = (x,y,z,1)^\top$ is the homogeneous coordinates in the world space.

By parameterizing the ray as the intersection of two orthogonal planes, $\boldsymbol{h}_u$ and $\boldsymbol{h}_v$, the ray-splat intersection in Gaussian plane can be determined by solving:
\begin{equation}
[\boldsymbol{h}_u,\boldsymbol{h}_v]^\top  \cdot \mathbf{P}(u,v)=0,
\end{equation}

\begin{figure}[ht] 
\includegraphics[width=\linewidth]{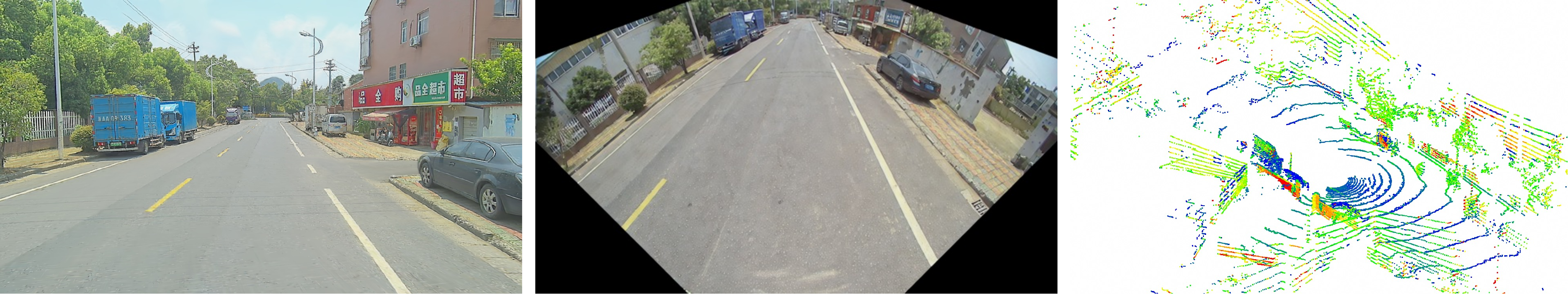}
\caption{We use a unified projection scheme to consistently process pinhole frames (left), fisheye frames (middle), and range-view LiDAR frames (right).}
\label{fig:projection} 
\end{figure}

Therefore, we can unify the projection scheme by changing $[\boldsymbol{h}_u,\boldsymbol{h}_v]$ for different projection models as shown in Fig.~\ref{fig:projection}. Specifically, for pinhole cameras~\cite{Kerbl20233dgs}, when given an image coordinate $(p_x,p_y)$, the definition of orthogonal planes is as follows:
\begin{equation}
\boldsymbol{h}_u=(-1,0,0,p_x)^\top,
\boldsymbol{h}_v=(0,-1,0,p_y)^\top
\end{equation}

For fish-eye cameras (cylindrical projection), when given an image coordinate $(p_x,p_y)$, the elevation angle can be obtained as $\phi=(p_x-c_x)/f_x$, where $c_x$ and $f_x$ represent the camera intrinsics. The definition of orthogonal planes is as follows,
\begin{equation}
\boldsymbol{h}_u=(\cos\phi,0,-\sin\phi,0)^\top, \\
\boldsymbol{h}_v=(0,-1,0,p_y)^\top
\end{equation}
%%% \boldsymbol{h}_v=(0,-1,0,(p_y-c_y)/f_y)^\top

For LiDAR sensors, as our previous work~\cite{chen2024lidar}, when given an azimuth angle $\theta$ and elevation angle $\phi$, the definition of orthogonal planes is as follows: 
\begin{equation}
\begin{aligned}
\boldsymbol{h}_u&=(\sin\phi,-\cos\phi,0,0)^\top, \\
\boldsymbol{h}_v&=(\sin\theta\cos\phi, \sin\theta\sin\phi,-\cos\theta,0)^\top.
\end{aligned}
\end{equation}

\textbf{Support Large-scale Scenes.} 
The Vanilla 3DGS~\cite{Kerbl20233dgs} does not support large-scale scene reconstruction due to limited graphics memory, long optimization time and noticeable appearance variations. While several recent methods~\cite{kerbl2024h3dgs,ren2024octree} propose Level-of-Detail (LoD) representations, we find distributed reconstruction sufficient and it reduces extra training on hierarchical assets. Specifically, to support large-scale scenes training, we cluster sensor frames through their XOY-location into $40 \times 40$ square-meters zones, using $25\%$ relative padding to compensate view-direction related issues with divide-and-conquer approach. During rendering, we use two threads for double-buffering, swapping-in related Gaussians when the vehicle is moving between two blocks.

\textbf{Paralleled training.} Inspired by Grendel-GS~\cite{zhao2024scaling} which parallelizes the training of distributed 3D Gaussians, we propose the following modifications to adapt the method to neural Gaussians: (1) We introduce an adaptive strategy based on rendering time to allocate continuous image regions across multiple GPUs, enabling pixel-level parallelism. (2) On each GPU, we dynamically adjust the Gaussian parameters and the associated MLP parameters for color and opacity, integrating the MLP parameters into the load balancing strategy to optimize resource utilization. (3) We employ a sparse all-to-all communication scheme to transfer the required Gaussian data to corresponding pixel partitions across GPUs. Finally, to ensure compatibility with end-to-end simulation systems during real-time rendering on one GPU, we seamlessly concatenate the MLP and Gaussian parameters trained on different GPUs before rendering, regardless of the original distribution order of the model.

\subsection{Scene Editing and Expansion}
\label{sec:method:edit}

\textbf{Gaussian Inpainting.} Given reconstructed scenes with dynamic and static foreground actors, our first step is to seamlessly remove unnecessary actors before insertion, expansion, and harmonization. In this step, we extract undesired foreground regions and substitute nearby affected anchor features -- mostly due to shadows, occlusions and incomplete scans -- in a patch-matching style~\cite{barnes2009Patchmatch}. We define the following affinity score $\mathcal{A}$, to sort and pick up the best source patch $\mathcal{S}$ searched from lateral and longitudinal directions to replace the original patch $\mathcal{T}$:
\begin{equation}
\mathcal{A} = \mathrm{softmax}([\mathbf{F}_{\mathcal{T}}\oplus \mathrm{PE}(\mathbf{L}_{\mathcal{T}})][\mathbf{F}_{\mathcal{S}}\oplus \mathrm{PE}(\mathbf{L}_{\mathcal{S}})]^\top),
\label{equ:affinity}
\end{equation}
where for the anchor feature vectors in each patch $\mathcal{X} \in \{ \mathcal{S}, \mathcal{T} \}$, we extract and aggregate them to obtain the corresponding anchor feature matrix $\mathbf{F}_{\mathcal{X}}$, and concatenate position encoding $\mathrm{PE}(\mathbf{L}_\mathcal{X})$ for the location $\mathbf{L}_\mathcal{X}$ of each contained anchor. We also measure the appearance similarity according to the nearest-neighbor distance field~\cite{barnes2009Patchmatch} between two patches as additional comparison term. We hierarchically use three levels of patch size as $8^2 \rightarrow 4^2 \rightarrow 2^2$ square meters to replace with candidate source patches in a coarse-to-fine strategy gradually.

\begin{figure}[htbp] 
\includegraphics[width=\linewidth]{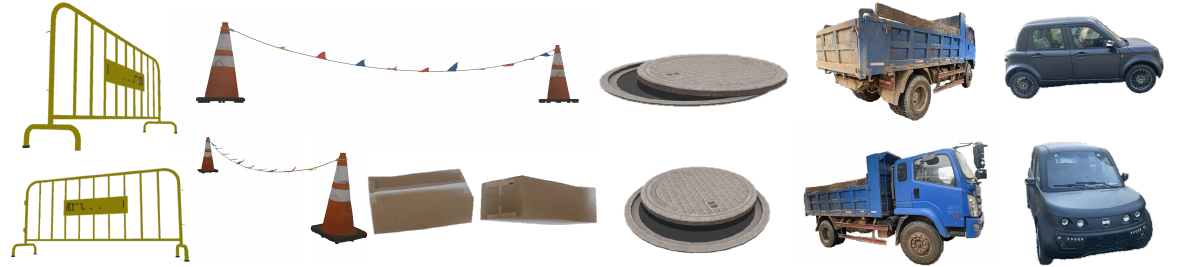}
\caption{Representative GS objects rendered from different perspectives.}
\label{fig:objects} 
\end{figure}

\textbf{Establishing Model Database.} We acquire high-quality actors represented in neural Gaussians through surround-view reconstruction. For general objects downloaded or manufactured as textured mesh, we use render engines to relight them and shutter surround views. For dynamic vehicles, we use the public dataset 3DRealCar~\cite{du20243drealcar}, which already scanned thousands of road actors in a similar scheme. 
During training, we use image masks from different views to restrict the primitives included in the model. The average training time and PSNR for all models are 10 minutes and 27.9 respectively. Our model database currently includes 143 cars and 47 generic objects with some assets shown in Fig.~\ref{fig:objects}.

\begin{figure}[ht] 
\includegraphics[width=\linewidth]{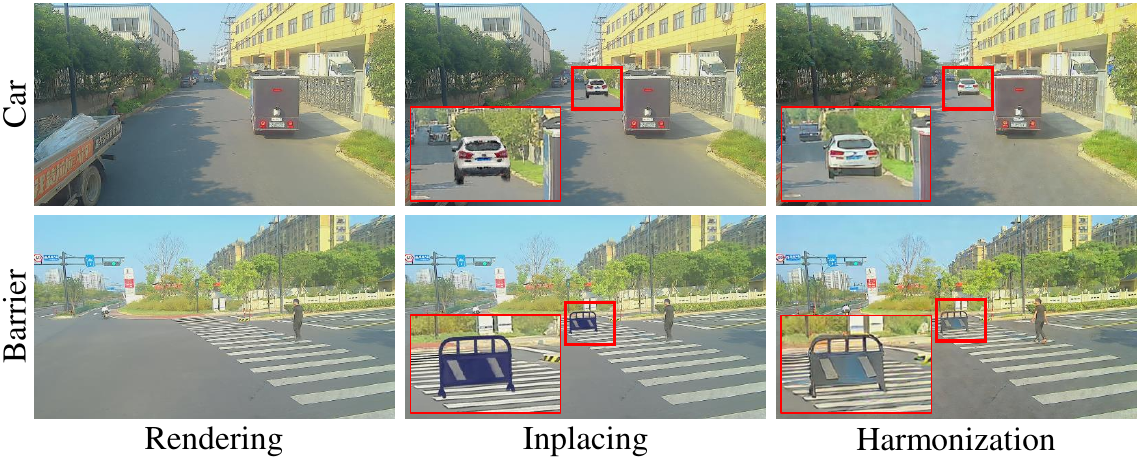}
\caption{Qualitative comparisons of dragging-in reconstructed models (middle) and using a camera diffusion model to refine their harmony (right).}
\label{fig:editing_workflow} 
\end{figure}

\textbf{Gaussian Inplacing.} We then inject these foreground actors into the reconstructed scene if necessary. We build a 3D interactive editing tool to manually specify the initial model matrix of each actor, and the trajectory for each moveable object. The 3D oriented bounding box of each actor is adjointly transformed to generate ground truth annotations. As model scans and road scans may vary in weather and lighting conditions, we optionally use controllable diffusion models explained below for obtaining seamless insertion. Fig.~\ref{fig:editing_workflow} shows several effects of our object insertion and diffusion-based harmonization.

\textbf{Controllable Diffusion Models for Expansion.}
Given single-pass scans, we leverage Stable Diffusion v1.5~\cite{rombach2022high} with ControlNet~\cite{zhang2023adding} and LiDM~\cite{ran2024LiDM} for camera and LiDAR frames, respectively, to generate extra frames for scene expansion. Specifically, we use driving sequences across different road sections and weather conditions to construct the training dataset following the strategy in FreeSim~\cite{fan2024freesim}.
During training, we keep the Variational AutoEncoders (VAE) parameters frozen and train the UNet and ControlNet for 50k iterations.
During expansion, we input GS rendering results on the required rendering pose to these diffusion models, with the diffusion process to optimize them as:
\begin{equation}
\mathcal{L} = \mathrm{E}_{\boldsymbol{z}_0,c,t,\epsilon}(\left\|\epsilon_{t}-\epsilon_{\theta},(\boldsymbol{z}_t,c,t)\right\|_2^2),
\end{equation}
where $\epsilon_{t}$ represents the random noise at time step $t$, $\epsilon_{\theta}$ denotes our trainable denoising model, $\boldsymbol{z}_t$ refers to the noisy latent, $\boldsymbol{z}_0$ represents the latent of the ground truth image from the VAE, and $c$ is the latent of image condition from GS rendering results. Then, during reconstruction, we add the diffused image for training GS assets to gradually enlarge the range of scenes.

\subsection{Ablation Studies}
\label{sec:method:ablations}

We quantitatively evaluate our methods and strategies against other applicable choices on the Para-Lane~\cite{ni2025paralane} dataset including 45 driving clips with 150 meters average cumulative distance, and report geometric, appearance, and efficiency metrics in Tab.~\ref{tab:exp_recon}.
All methods use 30,000 iterations during training, and the starting and stopping iterations of Gaussian densification are 500 and 15,000, respectively.
For camera reconstruction, our chosen neural Gaussians representation reaches better rendering quality than the vanilla 2DGS~\cite{huang20242dgs}, but greatly outperforms a candidate NeRF~\cite{yang2023emernerf} method on the rendering FPS. For LiDAR reconstruction, we additionally achieve better training speed than a state-of-the-art LiDAR-NeRF method LiDAR4D~\cite{zheng2024lidar4d}. All experiments in this paper are conducted on single or multiple NVIDIA RTX 3090s.

\begin{table}[htbp]
\caption{Performance of different representations. PSNR/SSIM for color of cameras and intensity of LiDAR, respectively.}
\centering
\fontsize{8pt}{9.6pt}\selectfont
\setlength{\tabcolsep}{5pt}
\begin{tabular}{l|cc|cc|c} 
\toprule
Method & CD$\downarrow$ & F-scr$\uparrow$  & PSNR$\uparrow$  & SSIM$\uparrow$  & FPS (Hz)$\uparrow$ \\ 
\midrule
EmerNeRF        & \multicolumn{2}{c|}{\multirow{4}{*}{\makecell{\color{gray} Not valid \\ \color{gray} for cameras}}}
                                                    & 24.97 & 0.668 & 0.2 \\
Scaffold-GS     & \multicolumn{2}{c|}{}             & 25.52 & 0.673 & 63.7 \\
2DGS            & \multicolumn{2}{c|}{}             & 24.28 & 0.681 & 29.9 \\
Ours            & \multicolumn{2}{c|}{}             & 26.82 & 0.685 & 46.7 \\
         \midrule
LiDAR4D & 0.201 & 0.916 & 33.12 & 0.914 & 1.6\\
Ours & 0.194 & 0.917 & 32.90 & 0.933 & 15.0\\
\bottomrule
\end{tabular}
\label{tab:exp_recon}
\end{table}

For long-term driving sequences, we compare several publicly available methods including Hierarhicical-3DGS~\cite{kerbl2024h3dgs} and Octree-GS~\cite{ren2024octree} also on the Para-Lane~\cite{ni2025paralane} dataset, to test the effectiveness of large-scale reconstruction. 
Specifically, we follow the preprocessing of Hierarchical 3DGS~\cite{kerbl2024h3dgs} and obtain deep supervision for optimization through the Depth-Anything-V2 model~\cite{yang2025depthanything}. The voxel size of our method is 8 centimeters, and the detail levels of Octree-GS is 7, corresponding to 0.2 centimeters and 17 centimeters of finest and coarsest voxels respectively.
As shown in Tab.~\ref{tab:large_scale}, by dividing into chunks properly, our method attains the highest image quality due to the finer spatial scale and smaller data size. Octree-GS achieves the fastest rendering speed with fewer primitives due to the LoD strategy, but this influences photo-realism on details.

\begin{table}[htbp]
\caption{Reconstruction performance on long sequences.}
\centering
\fontsize{8pt}{9.6pt}\selectfont
\setlength{\tabcolsep}{5pt}
\begin{tabular}{l|c|cc}
\toprule
Method                         & PSNR$\uparrow$ & Train (min)$\downarrow$ & FPS (Hz)$\uparrow$    \\
\midrule
Hierarchical 3DGS              & 26.35 & 161 & 27.7 \\
Octree-GS                      & 24.91 & 64  & 83.2 \\
Ours (w/o chunk-split)         & 25.05 & 187 & 32.4 \\
Ours                           & 26.82 & 132 & 46.7 \\
\bottomrule
\end{tabular}
\label{tab:large_scale}
\end{table}

Meanwhile, leveraging our proposed parallel training strategy significantly reduces the training time of Gaussians, reaching $65.5\% \sim 91.5\%$ parallelization w.r.t. different batch sizes given $4\times$ computing resources as summarized in Tab.~\ref{tab:parallel}.

\begin{table}[htbp]
\caption{Ablations of using paralleled training strategy.}
\centering
\fontsize{8pt}{9.6pt}\selectfont
\setlength{\tabcolsep}{6pt}
\begin{tabular}{l|c|c|c}
\toprule
Configuration   & Total Points & PSNR$\uparrow$  & Train (min)$\downarrow$ \\
\midrule
GPUs = 1, Bs = 1 & 318,385       & 26.82 & 132     \\
GPUs = 4, Bs = 1 & 325,807       & 26.71 & 68      \\
GPUs = 4, Bs = 4 & 346,519       & 26.79 & 44      \\
\bottomrule
\end{tabular}
\label{tab:parallel}
\end{table}

\begin{figure}[htbp] 
\includegraphics[width=\linewidth]{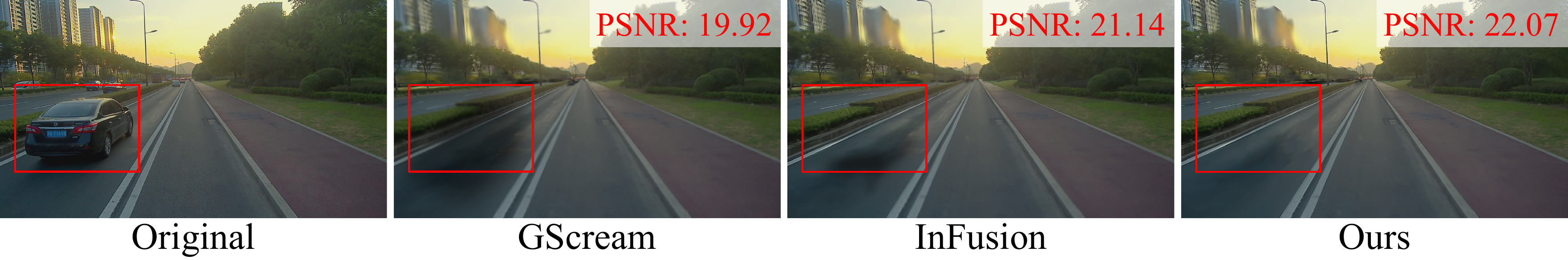}
\caption{Qualitative comparisons of different GS inpainting methods. We report the average PSNR value (red) of each method (GScream, InFuion and the proposed patch-based method) on six sequences with dynamic actors removed.}
\label{fig:inpainting} 
\end{figure}

We evaluate the performance of our patch-based GS inpainting by comparing it to two applicable methods as listed in Fig.~\ref{fig:inpainting}. GScream~\cite{wang2025gscream} is also based on the Scaffold-GS, using multi-view 2D observations as prior terms for optimizing 3D Gaussian attributes. It performs cross-attention between the target anchor and its surrounding anchors, guided by a 2D inpainted prior, to generate a more harmonious editing area. InFusion~\cite{liu2024infusion} combines 2D depth completion and 2D color inpainting to execute an additional round of 3D Gaussians re-training. In contrast, our method performs explicit cut-and-paste operations, avoiding a tightly coupled optimization or another round of re-training for flexibility. This strategy highly leverages the repetitive characteristic of urban roads, achieving comparable quality to these candidate methods.

We also evaluate the effectiveness of our scene expansion workflow leveraging diffusion models as shown in Tab.~\ref{tab:diffusion_study} on three sequences in the Para-Lane~\cite{ni2025paralane} dataset, with ground truth images in different lanes. We use two PSNR metrics to evaluate the lateral cross-lane rendering performance: across one lane (PSNR-1) and two lanes (PSNR-2). The average lateral shifting distance are 3.5 and 7 meters respectively. 
In conclusion, expanding the scene via stable diffusion can make the entire rendering results closer to the laterally shifted real scans, and coupling these diffused images into the reconstruction can maintain the original rendering speed with an acceptable trade-off on quality, avoiding re-generating through heavy models on new free-viewpoints.

\begin{table}[htbp]
\caption{Fidelity and rendering speed of scene expansion.}
\centering
\fontsize{8pt}{9.6pt}\selectfont
\setlength{\tabcolsep}{5pt}
\begin{tabular}{l|cc|c}
\toprule
Method                      & PSNR-1$\uparrow$ & PSNR-2$\uparrow$ & FPS (Hz)$\uparrow$    \\
\midrule
Ours (Inplacing)               & 18.11 & 16.45 & 27.3 \\
Ours (Harmonization)           & 20.63 & 20.19 & 0.057 \\
Ours (Harm. + Reconstruction)  & 20.41 & 19.83 & 27.1\\
\bottomrule
\end{tabular}
\label{tab:diffusion_study}
\end{table}

\section{Applications of Sensor Simulator}
\label{sec:app}

\subsection{Gaining Onboard Perception Accuracy}
\label{sec:app:perception}

\textbf{Augmenting rare training data.} Our onboard perception module mainly follows the BEVFusion framework~\cite{liang2022bevfusion,liu2023bevfusion} to process multi-modal sensor inputs, including synchronized LiDAR and camera frames. After BEV feature maps are obtained through this backbone, we connect multiple heads such as the detection head CenterNet~\cite{yin2021center}, the online mapping head MapTR~\cite{liao2023maptrv2}, and the occupancy head SurroundOCC~\cite{wei2023surroundocc}. 
When trained with extra simulated scans, we balance real scans and simulated scans as 50\%:50\% for the training set of a specific category, keeping all validation scans gathered from real-world, and remaining hardware and parameter settings of the training procedure unchanged.

\begin{figure}[htbp] 
\includegraphics[width=\linewidth]{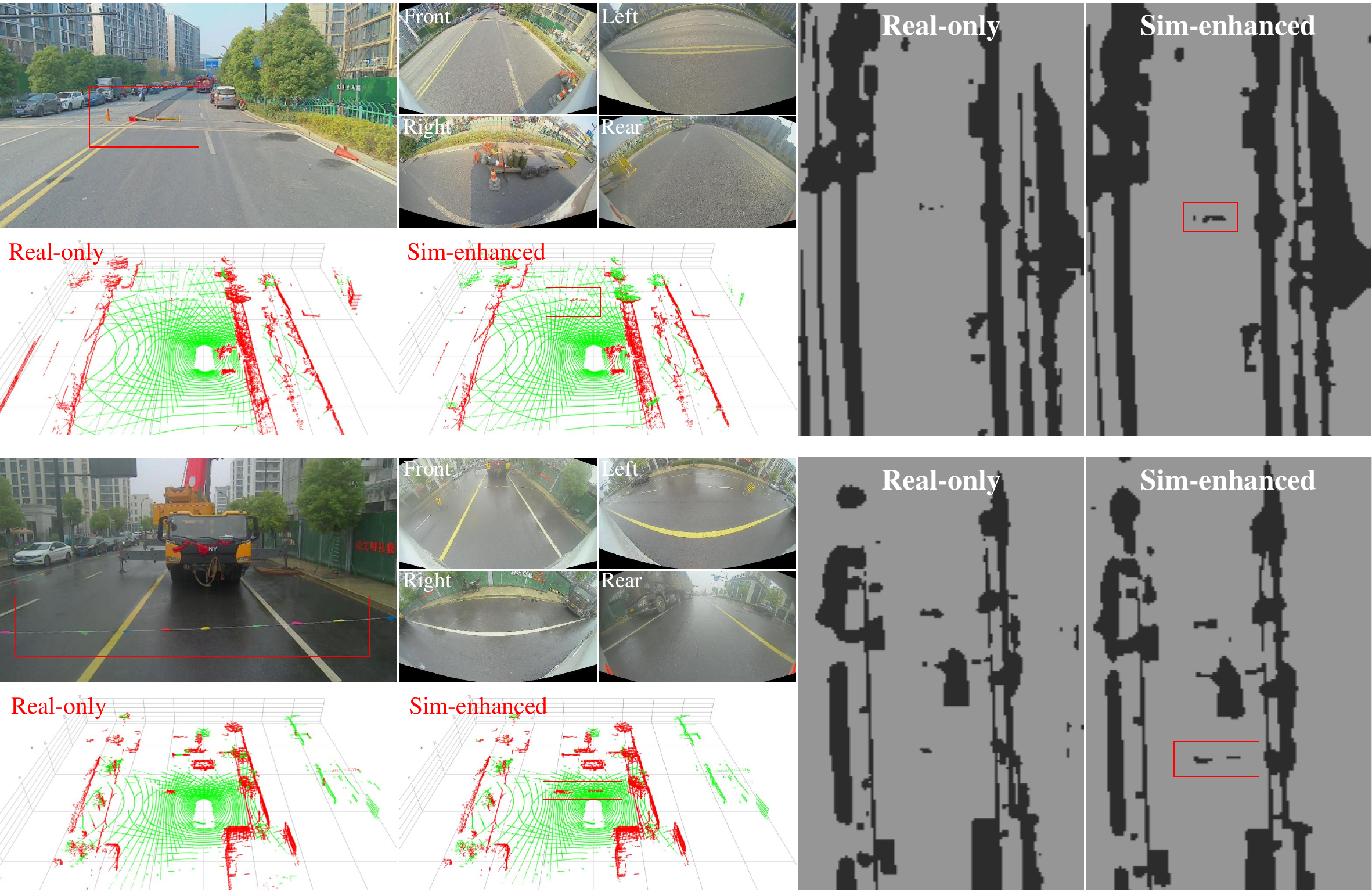}
\caption{We use simulated scans to augment our training set of rare objects and reserve real scans for the validation set, achieving better object detection recall on rare objects that are hard to collect and annotate.}
\label{fig:perception} 
\end{figure}

\textbf{Performance gain on the occupancy task.} Static foreground obstacles vary on urban roads and semi-enclosed areas where L4 autonomous vehicles drive, typically including hanging chains, traffic cones, express cartons, signboards, etc. Driving clips containing these objectives are rare and hard to be mined. Meanwhile, recognizing them through manual labeling involves high annotation costs, whereas generating driving clips from the model database saves the cost. We use our trained models (with or without simulated scans generated as Sec.~\ref{sec:method:edit}) to test and summarize the recall and precision per frame as listed in the left part of Tab.~\ref{tab:perception_performance} and Fig.~\ref{fig:perception}, reflecting the improvement of utilizing these trained data.

% \textbf{Evaluation on the OCC task.} We evaluate the occupancy prediction task using frame recall and frame prediction. We manually annotated corner cases (chains, traffic cones, boxes, signboards) region. A frame passes recall evaluation if any predicted occupied cell overlaps the annotation, yielding \(N_R^{TP}\) passing frames out of \(N_R\) total. For precision, scenes with point cloud noise, strong light, and fog were annotated for false detection-prone areas; a frame passes if no predicted cell overlaps these areas, counted as \(N_P^{TP}\) out of \(N_p\) frames.  Consequently, recall and precision are defined as \(N_R^{TP}/N_R\) and \(N_P^{TP}/N_P\), respectively.

\textbf{Performance gain on tracking reversing vehicles.} In urban roads where forwarding and stopped vehicles are common, clips with reversing vehicles are rare to collect, but they are dangerous and require a precise negative velocity estimation. Hence, we curate a training set having reversing vehicles, with half of the simulated clips edited through our scene editing workflow (Sec.~\ref{sec:method:edit}) from real-world clips. Since our onboard perception model is a multi-task architecture, we use the LoRA~\cite{hu2022lora} strategy to freeze those shared parameters and tune the detection head, reaching 5.5\% recall gain on reversing vehicles (R-Rec.) with the precision of general vehicle detection (A-Prec.) unchanged.

\begin{table}[ht]
\caption{Performance of occupancy and reversing vehicle prediction tasks (\%). Recall of statics are also separately listed.}
\fontsize{8pt}{9.6pt}\selectfont
\setlength{\tabcolsep}{2pt}
\centering
\begin{tabular}{l|ccccc|cc}
\toprule
       & \multicolumn{5}{c|}{Occupancy} & \multicolumn{2}{c}{Vehicle} \\
Method & Prec. & Rec. & R-Chain & R-Cone & R-Sign & A-Prec. & R-Rec. \\
\midrule
Ours (only Real) & 87.8 & 62.7 & 37.0 & 87.9 & 11.5 & 93.2 & 54.1 \\ 
Ours (with Sim)  & 86.5 & 70.4 & 68.6 & 91.9 & 30.8 & 93.4 & 59.6 \\
\bottomrule
\end{tabular}
\label{tab:perception_performance}
\end{table}

\subsection{Regressing End-to-End Algorithms}
\label{sec:app:end-to-end}

\textbf{Sim-to-Real Gap Evaluation.} To prove our effectiveness for validating end-to-end autonomous driving algorithms, we follow previous works discussing sim-to-real gaps~\cite{tan2023www} to test the cumulative drifts between using the same onboard algorithm processed on real-driven roads and re-processed on virtual environments. Since there are three simulators in the loop -- sensor, traffic, and dynamic -- where all of them could contribute to the final trajectory bias when offering inputs of perception, planning, and control tasks in the virtual environment regression, so we settle our baseline as skipping the onboard perception module and use transformed real perception results to feed into the rest simulators and algorithms. In contrast, our subject launches both the proposed sensor simulator and the onboard perception module to consist of a full-stack regression.

In our evaluation, we regress 50 clips of the following and cut-in scenarios with 110 meters and 180 meters average total distance, respectively, and compute the average cumulative drifting as listed in Tab.~\ref{tab:sim2real_gap}. The small difference of our cumulative drifting between the baseline and the subject w.r.t. the total distance reveals that we no longer require a transformation on perception results to skip virtual perceiving but can switch to such a GS-based real-time renderer to bring up the perception algorithm. We refer readers to our supplementary video for more perception and end-to-end demonstrations.

% \begin{table}[ht]
% \caption{Cumulative drifts reveal sim-to-real gap between our baseline and subject on whether perception with GS-based sensor simulator is launched. All units are in meters.}
% \fontsize{8pt}{9.6pt}\selectfont
% \setlength{\tabcolsep}{5pt}
% \centering
% \begin{tabular}{l|cccc|cccc}
% \toprule
%        & \multicolumn{4}{c|}{Lateral} & \multicolumn{4}{c}{Longitudinal} \\
% Method & Ave. & 90\textsuperscript{th} & 95\textsuperscript{th} & Max. & Ave. & 90\textsuperscript{th} & 95\textsuperscript{th} & Max.  \\
% \midrule
% Follow-Base & 0.03 & 0.08 & 0.08 & 0.09 & 0.69 & 0.81 & 0.95 & 1.10 \\ 
% Follow-Sub  & 0.16 & 0.43 & 0.47 & 0.63 & 1.38 & 3.33 & 3.85 & 4.60 \\ 
% \midrule
% Cut-in-Base  & 0.03 & 0.04 & 0.04 & 0.08 & 3.16 & 3.20 & 3.25 & 3.30 \\ 
% Cut-in-Sub   & 0.46 & 0.99 & 1.13 & 1.17 & 2.46 & 5.36 & 5.58 & 5.80 \\ 
% \bottomrule
% \end{tabular}
% \label{tab:sim2real_gap}
% \end{table}

\begin{table}[ht]
\caption{Cumulative drifts reveal a sim-to-real gap between our baseline and the subject of whether perception with GS-based sensor simulator is launched. All units are in meters.}
\fontsize{8pt}{9.6pt}\selectfont
\setlength{\tabcolsep}{4pt}
\centering
\begin{tabular}{l|cccc|cccc}
\toprule
       & \multicolumn{4}{c|}{Lateral} & \multicolumn{4}{c}{Longitudinal} \\
Method & Ave. & Med. & 80\textsuperscript{th} & 90\textsuperscript{th} & Ave. & Med. & 80\textsuperscript{th} & 90\textsuperscript{th}  \\
\midrule
Follow-Base & 0.03 & 0.03 & 0.06 & 0.08 & 0.69 & 0.70 & 0.80 & 0.81 \\ 
Follow-Sub  & 0.16 & 0.09 & 0.29 & 0.43 & 1.38 & 0.85 & 2.46 & 3.33  \\ 
\midrule
Cut-in-Base  & 0.03 & 0.02 & 0.03 & 0.04 & 3.16 & 3.19 & 3.19 & 3.20 \\ 
Cut-in-Sub   & 0.46 & 0.29 & 0.72 & 0.99 & 2.46 & 0.90 & 4.82 & 5.36 \\ 
\bottomrule
\end{tabular}
\label{tab:sim2real_gap}
\end{table}

\textbf{Scenario Generation and Behavior Validation.}
Following the approach in DriveFuzz~\cite{kim2022DriveFuzz}, we develop a fuzzing framework to generate and mutate driving scenarios. This framework converts the behaviors of actors from real driving clips into parameterized behavior trees for editable operations. It can also synthesize interactive behavior trees with diverse driving styles. When cooperating with a real-time sensor simulator, the fuzzing framework can efficiently bring up the perception algorithm and validate the sensor-to-control full-stack in reconstructed scenes with appropriately placed surrounding actors.

\begin{figure}[h!tbp]
  \centering
  \includegraphics[width=\linewidth]{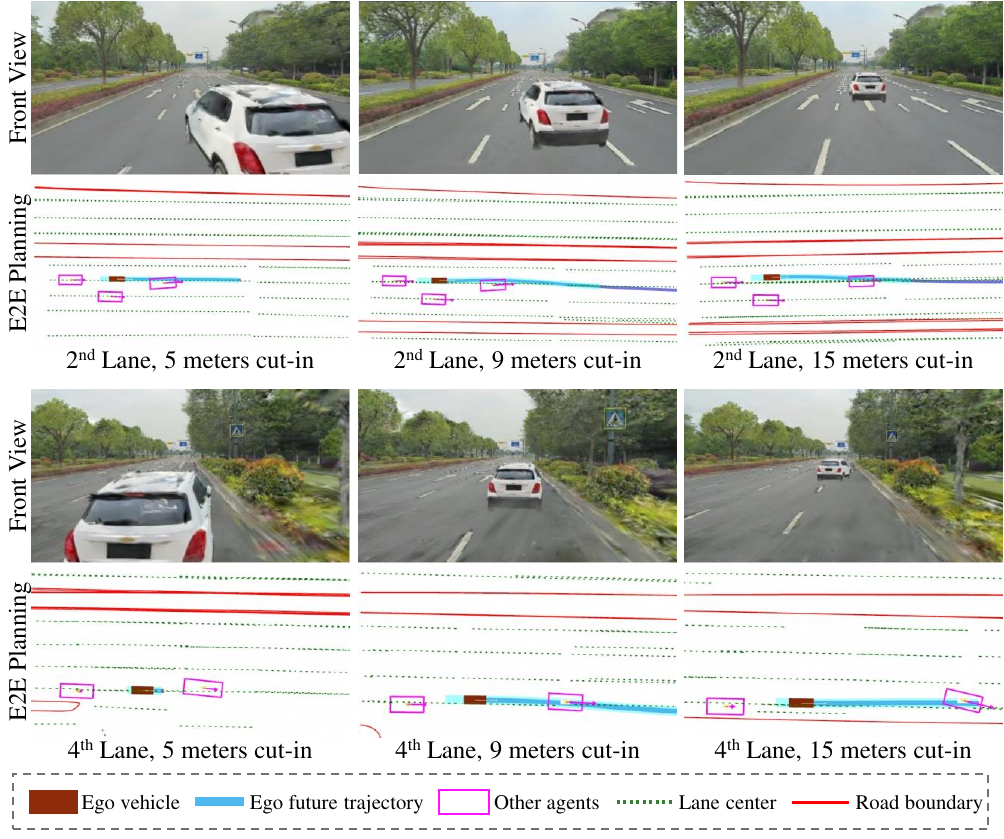}
  \caption{Regressing end-to-end algorithms on scenarios generated from a fuzzing framework.}
  \label{fig:interaction_loop}
\end{figure}

As a showcase, we can efficiently generate multiple virtual scenes for highly risky cut-in scenarios defined by generalization boundaries (e.g., relative velocity and distance). In these scenes, the cut-in vehicle is positioned at varying distances and speeds. Fig.~\ref{fig:interaction_loop} demonstrates our capability to launch and validate a sensor-to-planning end-to-end deep model~\cite{hu2023uniad}. In this model, surrounding vehicles actively try to cut into our lane at different speeds and distances, and our agent successfully makes diverse decisions, such as slowing down, ignoring, or changing lanes, as needed.

% To enhance the application of simulation in end-to-end autonomous driving algorithms, a scenario generalization framework termed the Fuzzing Engine is proposed to enable coverage-driven scenario generation and safety-critical scenario generation. Interactive behavior of critical agent from real-world data are translated into parameterized behavior tree representations for further editing. For cut-in scenarios, given specified behavioral generalization boundaries (e.g., relative velocity and distance thresholds), new interactive behavior trees with different driving style are automatically synthesized. These templates facilitate closed-loop simulations where AV interact with agents under safety-critical conditions.

% Generation toward high-risk scenarios is achieved through Bayesian optimization(BO) and genetic algorithms(GA), guided by evaluation metrics including collision and time-to-collision et al. Figure 4 illustrates visualizations of synthesized cut-in scenarios utilizing behavior tree-based agent editing and 3D gaussian splatting. This framework enhances safety validation for AVs through two mechanisms: 1) automated generation of corner case via parameter space exploration, and 2) data augmentation for end-to-end algorithm training via closed-loop interaction reconstruction. The framework demonstrates potential for improving the safety and performance of end-to-end systems through systematic data in adversarial scenarios.

\section{Conclusion}

In this paper, we introduced a Gaussian Splatting (GS)-based sensor simulation framework. By systematically refactoring critical components -- scene representation, sensor modeling, editing, and expansion -- we demonstrated that GS offers significant advantages over traditional NeRF-based methods: Our neural Gaussians representation, coupled with parallelized training strategies, achieves real-time rendering speeds while maintaining geometric and photometric fidelity. Meanwhile, the explicit nature of GS enables intuitive scene editing through drag-and-place operations and seamless integration of external assets, significantly reducing data collection and annotation costs. Furthermore, by harmonizing controllable diffusion models with GS, we expanded single-pass scans into physically compliant, free-viewpoint renderings, addressing challenges in crowd-sourced data incompleteness.

Applications in perception tasks, such as rare obstacle detection and reversing vehicle tracking, showed marked performance improvements when augmented with simulated data. Additionally, integrating our sensor simulator with traffic and dynamic simulators enabled full-stack validation of end-to-end autonomy algorithms, closing the sim-to-real gap and ensuring safe decision-making in high-risk scenarios.

Our future directions include extending the framework with prompt-based lightweight editing, enlarging the database in a self-evolved crowd-sourcing model collection, and computing precise ground manifolds for high-fidelity dynamic simulators

% \addtolength{\textheight}{-12cm}   % This command serves to balance the column lengths
                                  % on the last page of the document manually. It shortens
                                  % the textheight of the last page by a suitable amount.
                                  % This command does not take effect until the next page
                                  % so it should come on the page before the last. Make
                                  % sure that you do not shorten the textheight too much.

\bibliography{main}
\end{document}